\documentclass[review]{elsarticle}

\usepackage{lineno}
\usepackage[hidelinks]{hyperref}
\usepackage{graphicx}
\usepackage{caption}
\usepackage{tabularx}
	\newcolumntype{L}{>{\raggedright\arraybackslash}X}
\modulolinenumbers[5]
\usepackage{url}
\usepackage{enumitem}
\usepackage{algpseudocode}
\usepackage{pgfplots}
\usepackage{pgfplotstable}
\usepackage{amsmath}
\usepackage{booktabs}
\usepackage{cellspace}
\usepackage{siunitx}
\usepackage{setspace}
\usepackage[margin=1.5in]{geometry}
\usepackage{amsmath,amssymb,amsfonts, picins}
\usepackage{algorithm}
\usepackage[utf8]{inputenc}

\newcommand{\squishlist}{
 \begin{list}{$\bullet$}
  { \setlength{\itemsep}{0pt}
     \setlength{\parsep}{3pt}
     \setlength{\topsep}{3pt}
     \setlength{\partopsep}{0pt}
     \setlength{\leftmargin}{1.5em}
     \setlength{\labelwidth}{1em}
     \setlength{\labelsep}{0.5em}}
     }

\newcommand{\squishlisttwo}{
 \begin{list}{$\bullet$}
  { \setlength{\itemsep}{0pt}
     \setlength{\parsep}{0pt}
    \setlength{\topsep}{0pt}
    \setlength{\partopsep}{0pt}
    \setlength{\leftmargin}{2em}
    \setlength{\labelwidth}{1.5em}
    \setlength{\labelsep}{0.5em} } }

\newcommand{\squishend}{
\end{list}  }

\usepackage{comment}

\journal{Pattern Recognition} 

\begin{document}

\begin{frontmatter}

\title{TAA-GCN: A Temporally Aware Adaptive Graph Convolutional Network for Age Estimation}


\author[mymainaddress]{Matthew Korban}
\ead{acw6ze@virginia.edu}

\author[mysecondaryaddress]{Peter Youngs}
\ead{pay2n@virginia.edu}

\author[mymainaddress]{Scott T. Acton\corref{mycorrespondingauthor}}
\cortext[mycorrespondingauthor]{Corresponding author}
\ead{acton@virginia.edu}

\address[mymainaddress]{Department of Electrical and Computer Engineering, University of Virginia, Charlottesville, VA 22904}
\address[mysecondaryaddress]{Department of Curriculum, Instruction and Special Education, University of Virginia, VA 22904}

\begin{abstract}
This paper proposes a novel age estimation algorithm, the Temporally-Aware Adaptive Graph Convolutional Network (TAA-GCN). Using a new representation based on graphs, the TAA-GCN utilizes skeletal, posture, clothing, and facial information to enrich the feature set associated with various ages. Such a novel graph representation has several advantages: First, reduced sensitivity to facial expression and other appearance variances; Second, robustness to partial occlusion and non-frontal-planar viewpoint, which is commonplace in real-world applications such as video surveillance. The TAA-GCN employs two novel components, (1) the Temporal Memory Module (TMM) to compute temporal dependencies in age;
(2) Adaptive Graph Convolutional Layer (AGCL) to refine the graphs and accommodate the variance in appearance. The TAA-GCN outperforms the state-of-the-art methods on four public benchmarks, UTKFace, MORPHII, CACD, and FG-NET. Moreover, the TAA-GCN showed reliability in different camera viewpoints and reduced quality images.
\end{abstract}


\end{frontmatter}

\section{Introduction}
\label{sec:intro}

Age estimation has evolved from a carnival curiosity to an established task in computer vision \cite{mansouri2020survey}. It has many applications such as human-computer interaction (HCI), biometrics, age-restricted security control, video surveillance, and teacher-student differentiation in the classroom.
However, age estimation brings several major challenges, including the following: (1) the variance of appearance and facial expression, (2) viewpoint variations, and (3) non-ordinal temporal dependencies between ages. These challenges hinder the current standard methods to estimate age effectively. Therefore, we propose an algorithm including several novel components to handle the challenges above, which we explain as follows:

\subsection{Variance in Appearance}
People of the same age have remarkable variance in their appearance \cite{liu2019TIP}, which makes the age estimation challenging. To resolve this issue, some researchers have suggested age, gender and racial grouping \cite{liu2019TIP, sun2021info}. Nevertheless, such approaches fail when the grouping strategy is erroneous or when the same age groups are highly varying. 

Current age estimation methods use raw images commonly with Convolutional Neural Networks (CNN) \cite{mansouri2020survey}. While there have been significant advances in developing effective CNN architectures such as AlexNet and VGG, CNNs models are commonly used for object/subject classification. So, they might be less effective for age estimation because the features obtained from CNN models differ from those in the face. As opposed to objects/subjects, facial information are commonly centralized around specific facial keypoints. Moreover, in contrast to object/subject classes (such as a car, house, or pedestrian) that often differ in non-localized regions, the difference between age classes is mainly defined as local differences around specific facial keypoints. This fact is the same for the ages of similar classes.

In contrast to raw images, facial keypoints provide a more potent representation of the face, eliminating unnecessary data \cite{singh2017CVPR} and yielding critical information.
However, while some approaches used facial keypoints for age estimation \cite{kwon1993IRCV}, facial keypoints have not yet been effectively exploited in age estimation. It is because the current methods are based on 2D convolutional networks, which cannot effectively model the connectivity information in facial keypoints.
Specifically, 2D convolutional operators are applied on a fixed image grid where there might not be any explicit connections between facial keypoints on the 2D image space.

An appropriate way to model facial keypoints and their connectivity is by using graphs. The Graph Convolutional Network (GCN) has shown to be efficacious in solving graph-based problems such as action recognition using skeletons \cite{ korban2020ECCV}. Here, we propose a novel graph representation and a new GCN to model facial keypoints for age estimation effectively. To better accommodate the variance in appearance in facial graphs, we introduce an Adaptive Graph Convolutional Layer (AGCL) that adaptively refines the graphs.

\begin{figure}[!htbp]
	\centering
		\includegraphics[height=0.17\textheight]{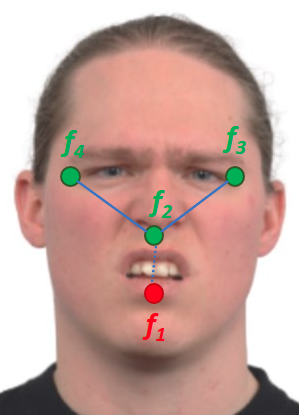}
  \includegraphics[height=0.17\textheight]{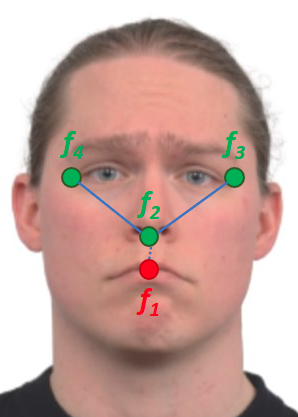}
  \includegraphics[height=0.17\textheight]{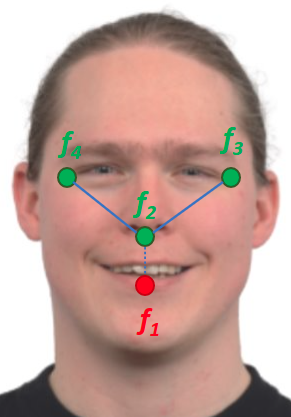}
  \includegraphics[height=0.17\textheight]{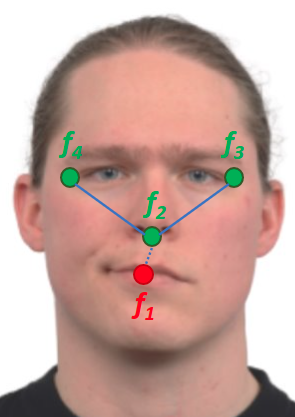}
	\caption{Four examples of simple facial graphs under different facial expressions. The appropriate graph nodes (green keypoints) are selected such that the features are less sensitive to the facial expression.} ~\label{Fig:expression}
\end{figure}

Another source of variance in the facial analysis is facial expressions that alter the structure of the face and facial keypoints.
There have been a few recent studies investigating age estimation under facial expressions. For example, \cite{lou2017PAMI} learned both the expressions and the ages jointly. Their method, however, requires complex learning and prior knowledge regarding facial expressions. We put forth a simple yet effective algorithm to make the GCN less sensitive to facial expressions, unlike this joint approach. An example of facial expression-insensitive graph can be seen in Fig. \ref{Fig:expression} where $f_2$, $f_3$, and $f_4$ almost remain in the same positions despite varying facial expressions that alters $f_1$ (face images are collected from  the RaFD dataset \cite{langner2010RaFD}). 

\vspace{10pt}

\subsection{Viewpoint Variations}
\begin{figure}[!htbp]
	\centering
		\includegraphics[height=0.22\textheight]{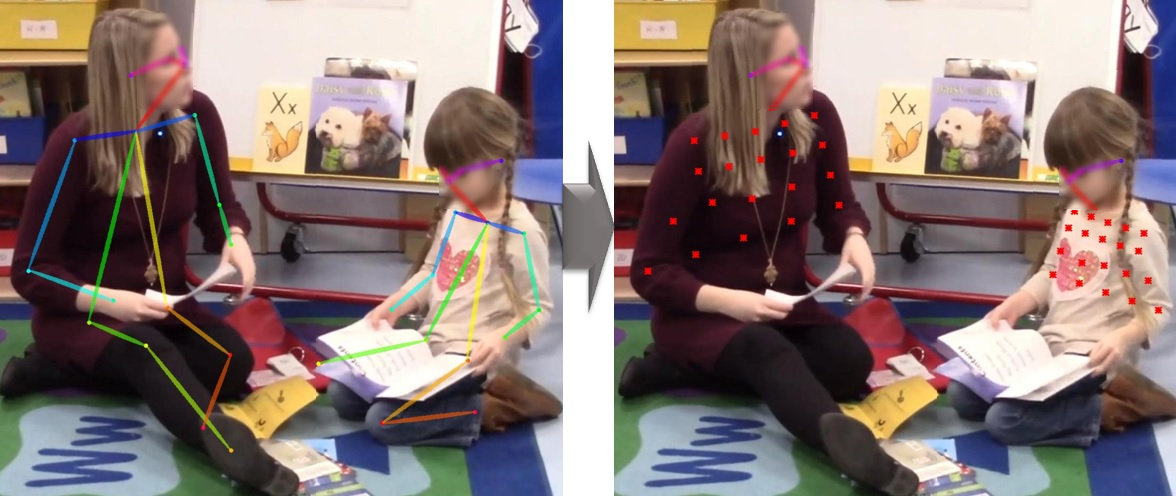}
	\caption{An example of a real-life scenario where skeleton and clothing information is more helpful than the face. In this frame sample from a classroom video, the teacher and student are captured from different viewpoints. As a result, their faces are only partially visible. However, their skeletons and clothing provide more critical information about their ages. To include this information, we propose to use Skeletal-Cosmetic (SC) keypoints (right image) which are obtained from detected skeletons (left image). ~\label{Fig:classroom}}
\end{figure}

In typical real-life scenarios, videos are captured from different camera angles and viewpoints. Most of the reports in the literature are heavily dependent on the single-view frontal face in which the face is frontal-planar with the imaging plane. A few existing strategies accommodate various camera angles in age estimation, including the estimation of various camera parameters \cite{nguyen2015sensor} and geometric parameters \cite{wu2012info}. However, these methods can handle only limited variations in viewpoints. In real-world applications such as video surveillance, the viewpoints are highly varying. 
As a result, a given face might be partially visible from certain viewpoints. An example is shown in Fig. \ref{Fig:classroom} in which a teacher's face and a student's face are partially visible in given viewpoints (Fig. \ref{Fig:classroom} is collected from a classroom video dataset belonging to the University of Virginia). In such examples, distinguishing teachers and students based on their ages has essential applications in some  ongoing research such as  teacher tracking and classroom activity recognition \cite{korban2021Asilomar}.  A standard age estimation algorithm encounters difficulty distinguishing their ages based on only facial information in these cases. Therefore, additional information is required to differentiate various ages when the viewpoint changes.  

\vspace{10pt}

\begin{figure}[!htbp]
	\centering
	\begin{tabular}{c c c c}
		\includegraphics[height=0.2\textheight]{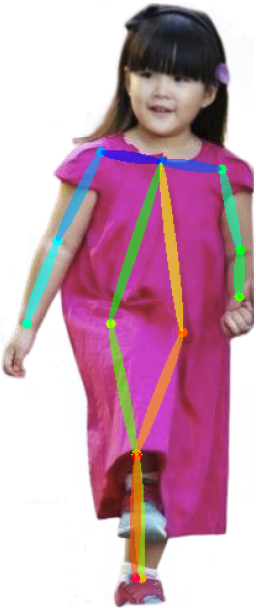}
		\includegraphics[height=0.29\textheight]{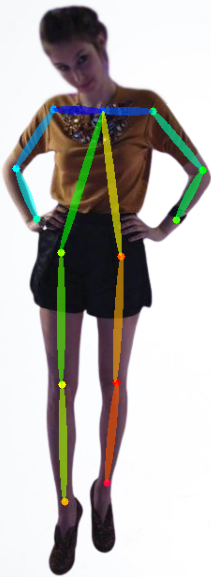}
		\includegraphics[height=0.3\textheight]{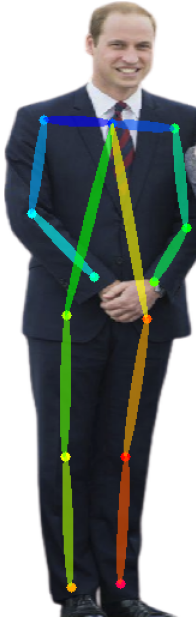}
		\includegraphics[height=0.3\textheight]{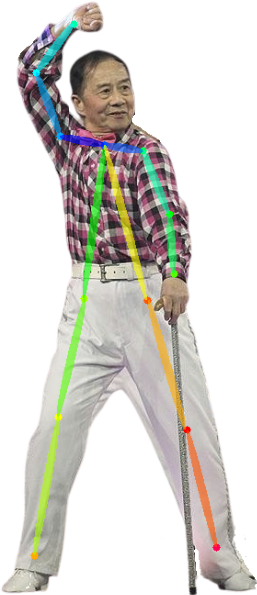}
			\end{tabular}
	\caption{Information beyond face is helpful for age estimation as people of different age groups have different skeleton structures, postures, and clothing.} ~\label{Fig:ages}
\end{figure}

People belonging to different age groups have different skeleton structures, adopt specific postures, and wear particular clothes; an example is shown in Fig. \ref{Fig:ages} (we collected the people's images from the Relative Human dataset \cite{sun2022CVPR}) and removed the background for more clarity). Although these cues are ``soft biometrics" rather than definitive features, such cues can be exploited in age estimation. The skeleton, posture, and clothing provide additional beneficial information in the attempt to distinguish different age groups, especially when the face is partially visible. As an example, in Fig. \ref{Fig:classroom}, the student and the teacher are now more distinguishable with their skeletal, posture and clothing information. (Fig \ref{Fig:classroom} is collected from a dataset belonging to the University of Virginia \cite{AIAI})
The combination of skeleton and clothing provides a unique and strong feature space distinguishing different age groups.
In this paper, for the first time, we propose the exploitation of the aforementioned additional soft biometrics, which we call \emph{Skeletal-Cosmetic} (SC) information, to improve the age estimation performance in varying viewpoints. To better handle the variance in SC graphs, our AGCL adaptively refines SC graphs separately from facial graphs 

\vspace{10pt}

\subsection{Non-ordinal Temporal Dependencies}
\label{Sec:TMM_intro}
\begin{figure}[!htbp]
\renewcommand{\tabcolsep}{0.5pt}
	\centering 
		\begin{tabular}{ccccc}
		\includegraphics[height=0.16\textheight]{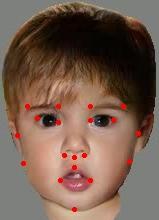} &
		\includegraphics[height=0.16\textheight]{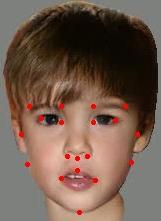} &
		\includegraphics[height=0.16\textheight]{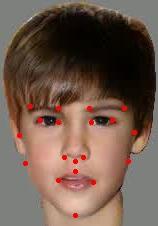} &
		\includegraphics[height=0.16\textheight]{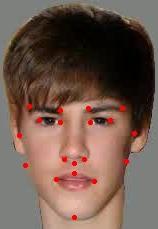} &
		\includegraphics[height=0.16\textheight]{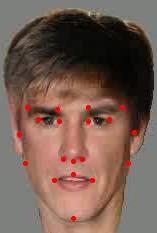} \\
		age = 1 & age = 11 & age = 21 & age = 31 & age = 41
    	\end{tabular}
	\caption{An image sequence exemplifying the temporal dependencies between neighboring ages. ~\label{Fig:temporal}}
\end{figure}

Aging is a temporal process, and as a result, neighboring ages are temporally dependent. An example of the aging process is shown in Fig. \ref{Fig:temporal}. Temporal dependencies help distinguish various age groups. However, these temporal dependencies are non-ordinal since age data samples are independent images from different individuals with no explicit temporal connections. In contrast, ordinal temporal sequences such as action video samples are typically continuous temporal frames showing the same individual in action. Therefore, standard temporal networks cannot capture such non-ordinal temporal dependencies among different ages.

Some strategies such as multi-stage classification \cite{yang2018IJCAI}, ranking \cite{chen2017CVPR}, and grouping \cite{tan2017CVPR} use multiple classifiers to implicitly exploit the temporal properties of ages at a high computational cost. Nevertheless, there does not exist a computationally efficient approach that takes advantage of temporal dependencies among different ages explicitly. \cite{dehshibi2019VC} proposes a kernel-based bi-directional PCA to find the kinship relationship between family members. However, this kinship relationship is limited to certain age groups (parent and child).  Moreover, the proposed learning process is not end-to-end and heavily depends on multiple stages including a pre-processing feature extraction using PCA.
To fill this gap, we put forth a new Temporal Memory Module (TMM) that captures non-ordinal temporal dependencies among a wide range of people with different ages. With a single-stage classification and end-to-end network, our proposed method is also computationally efficient.

The main \textbf{contributions} of this paper are:
\vspace{-5pt}
\begin{itemize}
    \item \textbf{We are the first} to introduce a graph representation for age estimation. Our new graph representation has some benefits over the previous 2D image-based approaches: First, it more effectively models facial keypoints and their connectivity; Second, it allows to exploit our new Adaptive Graph Convolutional Layer (AGCL) and facial keypoint selection algorithm to make the age estimation less sensitive to facial expressions and other appearance variances. 
    \vspace{-5pt}
    \item We propose a new Temporal-Aware Adaptive Graph Convolutional Network (TAA-GCNN) that includes a new Temporal Memory Module (TMM) to capture non-ordinal temporal dependencies among different ages. However, standard temporal networks cannot capture such non-ordinal temporal dependencies. \textbf{We are the first} to explicitly exploit non-ordinal temporal dependencies among different ages with a computationally efficient single-stage classification.   
    \vspace{-5pt}
    \item Our TAA-GCN also includes another new module, Adaptive Graph Convolutional Layer (AGCL), that refines the facial and SC graphs adaptively to improve the age estimation performance under the variance of appearance. 
    \vspace{-5pt}
    \item \textbf{We are the first} to include several soft biometrics including skeleton structure, posture and clothing (Skeletal-Cosmetic) information to improve the age features representation, especially in real-world applications where facial information is only partially available from certain viewpoints. 
    \vspace{-5pt}
    \item We conducted thorough experiments to estimate the age in-the-wild. 
    \vspace{-5pt}
    \item Our proposed method \textbf{outperforms} the state-of-the-art strategies as demonstrated on four public datasets, MORPHII, CACD, UTK-Face, and FG-NET.  
\end{itemize}

\section{Related Work}

\subsection{Age Estimation}
Classical approaches utilized hand-crafted features such as wrinkles, landmark-based features for age estimation. \cite{YHCVPR94} used natural wrinkles defined a search region on the face where facial wrinkles are more common. This early method could only estimate three age groups, baby and adult, and senior. \cite{ng2018IVC} overcame this limitation by estimation a variety of ages using Multiscale Wrinkle Patterns (MWP) features. Similar to natural wrinkles \cite{YHCVPR94}, MWP also are defined on multiple search regions. However, MWP included several other attributes such as shape and textures to enrich the age feature representation.
Some few research have been based on facial landmark features.  \cite{hsu2017CVPRW} extracted Component extracted Bio-Inspired Feature (BIF) from facial landmarks using pyramid of convolution filters. \cite{hammond2020AS} combines facial landmark points and gravity moment and builds a matrix that represents the the juvenile age range. 
Other features have been based on geometry, active shape, appearance \cite{mansouri2020survey}, and  relative-order information in different ages \cite{chang2015TIP}.
Different classifiers have been used with hand-crafted features such as Relevance Vector Machine (RVM) \cite{thukral2012ICASSP}, ratio matching\cite{YHCVPR94}, and Support Vector Regression \cite{ng2018IVC}.

With the recent advancement of deep learning, deep networks have been used for age estimation. Some researchers suggested improving the training stage. For instance, \cite{hu2016TIP} propose a CNN architecture to exploit age differences and reduce the number of training labels.
\cite{li2017PR} proposed a CNN architecture with a cumulative hidden layer and extracts discriminative aging features to resolve the issue of imbalance data. \cite{li2020PR} suggested a label refinery network (LRN) to refine the age labels for a more effective training. \cite{deng2021CVPR} suggested a Progressive Margin Loss (PML) to include the dependencies between the intra-class and inter-class variance in various age groups. 
Some researchers proposed to improve the age feature representation by revising the deep network architecture. \cite{taheri2019NC} suggested to use multi-scale output connections from different CNN layers to include diverse face features. \cite{tan2019IJCAI} suggested using multiple features extracted from local and global regions. \cite{wang2022TIP} proposed to weight important facial patches using Attention-based Dynamic Patch Fusion (ADPF).

Some researchers recommended assigning classifiers to various age groups. For example, \cite{yang2018IJCAI} suggested a multi-stage classification approach for different age groups. \cite{xie2020dTIFS}  enhanced the multi-group classification using Ordinal Ensemble Learning.
Another strategy for age estimation has been using additional human attributes or features. For instance, \cite{liu2019TIP} introduced an age grouping strategy including genders and sub-groups to facilitate the age estimation task.
\cite{sun2021info} proposed a deep conditional distribution learning which is conditioned to several attributes such as gender and age.

There have been a few approaches such as \cite{alonso2020BIOSIG} that tried to estimate age with partial information. To accomplish this goal, they model different face regions , such as eyes and nose separately. However, \cite{alonso2020BIOSIG} is still dependent on the high-quality frontal face and might not work in-the-wild scenarios when images are captured in reduced quality and from different camera angles. Our method, however, is reliable under various camera angles and for reduced quality images, as the skeleton, posture and clothing information are less affected than the face by camera viewpoints and image quality.

\subsection{Graph Convolutional Network}
Graph Convolutional Network (GCN) has been used in several computer vision tasks such as action recognition \cite{shi2022PR}, brain disorder prediction \cite{jiang2020CBM}, image retrieval \cite{chaudhuri2019CVIU}, person re-identification \cite{wu2022PR}, and recommendation systems \cite{yu2020TKDE}.
There are different types of GCN which have been introduced based on various applications. An Spatial GCN can encode spatial properties of data such as image pixels \cite{qin2018IGRSL}. A Temporal GCN computes the temporal dependencies of input like sequential traffic data \cite{zhao2019TITS}. A spatial-temporal GCN captures the information in both spatial and temporal domains such as pixels and sequential frames in activity videos \cite{ding2022EAAI}. 

For different tasks, the researchers designed various GCN architectures. For example, \cite{shi2022PR} designed a human pose-aware GCN to model the dependencies among human skeleton joints and body parts. \cite{jiang2020CBM} proposed a Hierarchical GCN to learn from different ROIs in fMRI data of the brain. \cite{chaudhuri2019CVIU} introduced a Siamese GCN to improve the discriminative property of image representations in image retrieval. \cite{wu2022PR} proposed a part-guided GCN to model the structural relationship in the learned features for person re-identification. \cite{yu2020TKDE} suggested an adversarial GCN to overcome the incomplete and noisy social network information for recommendation systems.  
According to our age estimation task, we designed a GCN which is (1) temporally aware of different age groups and (2) is adaptive to spatial-temporal variance in facial and SC graphs for different human faces, poses and clothing.

\section{Methodology}

\subsection{Overview and Terminology}
\label{Sec:overview}

\begin{figure*}[!htbp]
	\centering
		\includegraphics[width=1.0\textwidth]{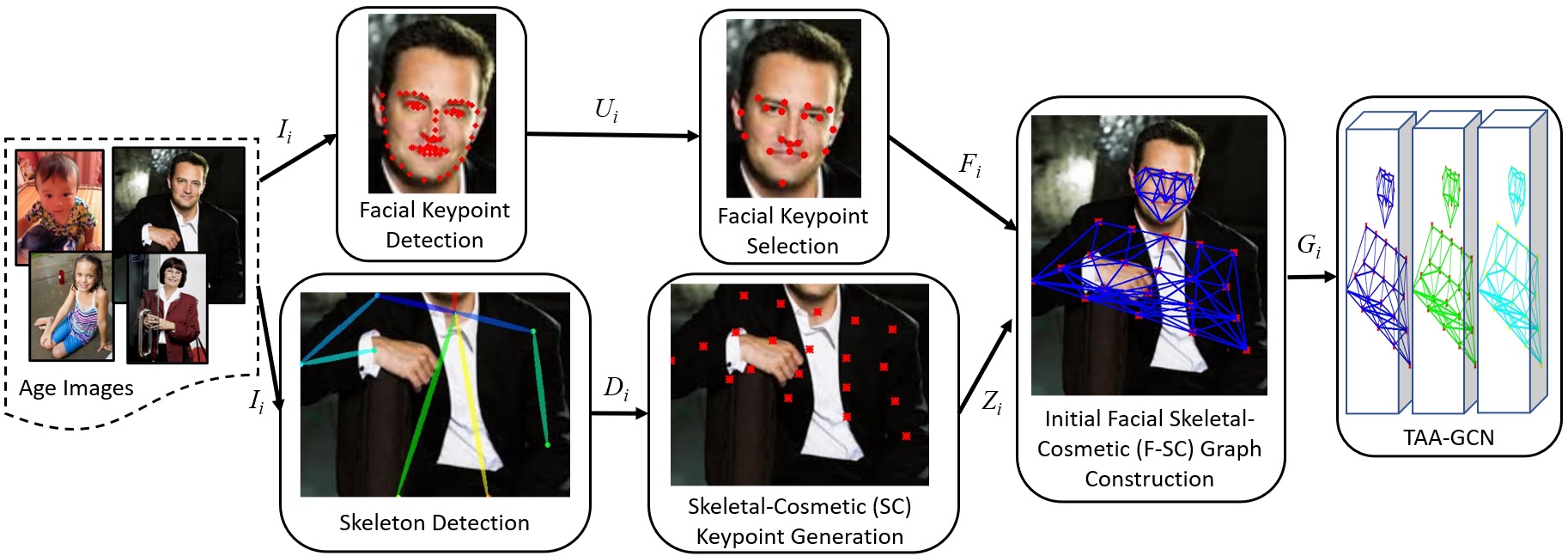}
	\caption{Overview of our proposed pipeline for age estimation. From a set of age images including $I_i$, we detect initial facial keypoints ($U_i$) and skeleton joints  ($D_i$) using the methods of \cite{bulat2017ICCV} and \cite{cao2019PAMI}, respectively.  Subsequently, a facial keypoint selection algorithm picks the most suitable subset of keypoints, $F_i$, that are more informative and less sensitive to facial expressions. Simultaneously, from  $D_i$ we generate the Skeletal-Cosmetic (SC) keypoints, $Z_i$, to include skeleton, posture, and clothing information. Then, given the previously obtained keypoints (and their corresponding patches of pixels), we construct an initial Facial Skeletal Cosmetic (F-SC) graph, $G_i$, by defining the initial connectivity (edges) among keypoints (nodes). Finally, our proposed Temporal-Aware Adaptive Graph Convolutional Network (TAA-GCN) estimates $G_i$'s age.  \label{Fig:overview}}
\end{figure*}

An overview of our proposed age estimation pipeline is shown in Fig. \ref{Fig:overview}. Given a set of images of people of varying age $I^S = \{I_i, i=1,2, \cdots, \Lambda\}$, for each $I_i$, we calculate a group of initial facial keypoints $U_i = \{u_k, k=1,2, \cdots N\}$ using \cite{bulat2017ICCV}, where $B$ is the number of samples in the dataset, and $N$ is the number of initial facial keypoints. Concurrently, for each $I_i$, we calculate a collection of skeleton joints $D_i = \{d_k, k=1,2, \cdots, M\}$ utilizing OpenPose \cite{cao2019PAMI}, where $M$ is the number detected joints. 
Next, we use our \emph{Facial Keypoint Selection} algorithm to select the most informative, yet least expression sensitive facial keypoint indices as $R=\{r_k, k=1,2, \cdots, N'\}$. We obtain $R$ offline and during the pre-training phase and use it as a fixed parameter to select more effective facial keypoints as $F_i = U_i(R)$ in the training and testing phases. 
On the other side, our \emph{Skeletal-Cosmetic Keypoint Generation} algorithm processes $D_i$ to generate SC keypoints $Z_i = \{z_k, k=1,2, \cdots, M'\}$ which represents both the skeleton and clothing ($M'$ here is the number of SC keypoints). For $F_i$ and $Z_i$ we create two sets of initial feature vectors of $X^F_i$ and $X^Z_i$, respectively.  
The full set of graph nodes is formed by concatenating the two sets of initial feature vectors as $V_i=\{X^F_i,X^Z_i\}$.

For facial and SC graph nodes (keypoints), we predefine initial adjacency matrices as $A^F$ and $A^Z$, respectively. $A^F$ and $A^Z$ indicate the connectivity information among the graph nodes in the face and body. Using the full adjacency matrix $A = \{A^F,A^Z\}$ (corresponding to both facial and SC graphs) and full set of nodes $V_i$, we construct an initial Facial Skeletal-Cosmetic (F-SC) graph as $G_i=\{V_i; E_i\}$, where $E_i$ is the set of graph edges and obtained from $A$. Finally, our TAA-GCN estimates $G_i$'s age $q_i$. Our TAA-GCN includes two new modules to improve the age estimation performance. First, AGCL to refine $G_i$ adaptively. Second, TMM to capture non-ordinal temporal dependencies among various ages. We explain all the terminologies and definitions with the corresponding reference sections of this manuscript in Table \ref{Tab:terms}

\begin{table}[h!tbp]
	\centering
	\caption{The terminologies used in this paper with descriptions and corresponding sections.}\label{Tab:terms} 
\resizebox{\textwidth}{!}{\begin{tabular}{ccc}
\hline
Terminology symbol & Description & Section\\ \hline
$I^S$ & set of images $I_i$ in the dataset, $0 \leq i \leq \Lambda$ & \ref{Sec:overview} \\  
$F_i$, $R$ & selected facial keypoints and their indices & \ref{Sec:FKS} , \ref{Sec:overview} \\
$U_i$ & initial set of facial keypoints $u_k$ before selection & \ref{Sec:FKS} \\
$K$, $\beta$ & number of neighboring keypoints, and neighboring nodes set & \ref{Sec:FKS} \\ 
$\zeta_k$ & relative distance in a facial neighboring keypoints & \ref{Sec:FKS}\\
$H$ & facial neighboring keypoints normalization term & \ref{Sec:FKS} \\
$v_k^E$. $v_k^A$  & facial expression, and age variance & \ref{Sec:FKS}\\
$\eta$  & keypoint selection weighting parameter & \ref{Sec:FKS} \\
$N$, $M$ & number of initial facial keypoints and skeleton joints  & \ref{Sec:FKS}, \ref{Sec:SCKG}\\
$N'$. $M'$ & number of selected facial, and generated SC keypoints &  \ref{Sec:FKS}, \ref{Sec:SCKG}\\
$D_i$ & set of skeleton joints $d_k$ & \ref{Sec:SCKG}\\
$O$, $J$ & number of hierarchical levels, and graph nodes &  \ref{Sec:SCKG} \\
$X_i^F$ & set of initial facial feature vectors $x_i^F$ & \ref{Sec:SCGC} \\
$X_i^Z$ & set of initial SC feature vectors $x_i^Z$  & \ref{Sec:SCGC}\\
$A^F$, $A^Z$  & facial, and SC adjacency matrix  & \ref{Sec:SCGC}\\
$G_i$ , $V_i$, $E_i$  & F-SC graph, nodes, and edges & \ref{Sec:SCGC} \\
$e$ , $\rho$  & nodes edge, and correlation value & \ref{Sec:SCGC}\\ 
$P_{spt}$, $P_{temp}$ & Spatial, and Temporal age probability & \ref{Sec:TAAGCN}\\
$Q$, $a$ & number of ages, and age label & \ref{Sec:TAAGCN} \\
$W^F$, $W^Z$, & adaptive facial, and SC graph edge weights & \ref{Sec:TAAGCN}\\
$\phi$, $psi$ & adaptive activation function, and age weighting parameter & \ref{Sec:TAAGCN} \\
\hline
\end{tabular}}
\end{table}

\subsection{Facial Keypoint Selection}
\label{Sec:FKS}
Our new graph representation of the face allows us to select the most informative facial keypoints. Given the initial detected facial keypoints ($U_i$), our facial keypoint selection algorithm picks the keypoints ($F_i$) that are less sensitive to facial expressions while more sensitive to aging. We observed that relative distances between neighboring keypoints often change noticeably under different facial expressions. An example is shown in Fig. \ref{Fig:expression}, where the relative Euclidean distances between two keypoints $d(f_1, f_2)$ change during the expression ``smiling". In the aging process, however, the global positions of keypoints shift more dramatically. We exploit this fact to select the most effective facial keypoints based on the facial expression and aging variances.
We selected the most effective facial keypoints based on the images captured in-the-wild such as those for the UTKFace dataset \cite{zhifei2017cvpr}. We also further evaluated our method based on classical facial expressions on the PAL dataset \cite{minear2004PAL}.
Our facial keypoint selection algorithm aims to find the facial keypoints that are less sensitive to facial expressions while being informative enough to improve the age estimation performance. We designed such a trade-off in a data-driven way based on the datasets we used. The datasets include a variety of different individuals from different cultures and various ages.

\emph{Facial Expression Variance.}
First, for each facial keypoint $f_k$, we define a set of neighboring keypoints as $\beta(f_k) = \{f_n, n =0, 1 \cdots K\}$, where $K$ is the number of neighboring keypoints. Then, we calculate the sum of the relative Euclidean distance ($d$) of each keypoint $f_k$ to its neighbors as:

\begin{equation}
     \zeta_k = \frac{1}{H}\sum_{n=1}^{K} d(f_k, f_n)
\end{equation}

In the above,  $H = \underset{1 \leq i \leq N}{\max}(\zeta_i)$ is a normalization term, $f_n \in \beta(f_k)$, where $N$ is the number of keypoints. For each $f_k$, we compute the facial expression variance $v^{E}_k$ across all the samples in the dataset as:

\begin{equation}
 v^{E}_k  = \frac{1}{N} \sum_{k=0}^{\Lambda} (\zeta_k  - \frac{1}{N} \sum_{k=0}^{\Lambda} \zeta_k )^2
\end{equation}

\emph{Age Variance.}
We define the global distance of $f_k$ with respect to the root keypoint $f_R$ as $\gamma_k = d(f_k, f_R$). For each $f_k$, we calculate the age variance $v^{A}_k$ across all the images in the dataset:

\begin{equation}
 v^{A}_k  = \frac{1}{N} \sum_{k=0}^{\Lambda} (\gamma_k  - \frac{1}{N} \sum_{k=0}^{\Lambda} \gamma_k )^2
\end{equation}

\emph{Keypoint Selection.}
We select the facial keypoints with the lowest facial expression variances and highest age variances. Specifically, among the keypoints with the indices of $k \in N$, we select top-$N'$ keypoints with the highest $v^T_k$:

\begin{equation}
     v^T_k = \eta \cdot v^{A}_k + (1-\eta) \cdot (1-v^{E}_k) 
\end{equation}

In the above, $\eta$ is a weighting parameter. We store the selected keypoint indices $R=\{r_k, k=1,2, \cdots, N'\}$ to pick the selected facial keypoints $F_i = U_i (R)$ during the training and testing phases.  
The value for $\eta$ is obtained experimentally. Specifically, we first select uniform sampling values from the interval [0, 1] and narrow them down to the smaller interval that maximizes the overall age estimation performance. 

\subsection{Skeletal-Cosmetic (SC) Keypoint Generation}
\label{Sec:SCKG}
Pose estimation algorithms such as OpenPose \cite{cao2019PAMI} extract spatially consistent and stable keypoints which have been used in many applications such as action recognition using skeletal graphs \cite{korban2020ECCV}. Hence, we use the detected joints by the OpenPose as a reliable backbone to generate SC keypoints that represent skeletons, postures, and clothing. Given the initially detected skeleton joints $D_i$, we generate the SC keypoints, $Z_i$, by interpolating between $d_k \in D_i$. The main challenge here is that the number of detected joints varies based on the visibility of persons in different images. Therefore, different Hierarchical Levels (HL) can be detected for each image. We define a HL with $O$ levels, as a set of joints that have a similar parent-child ranking in a human body skeleton. For example two shoulders or arms are in the same HL. Within the same HL, the number of detected joints can be also vary for different image samples according to the camera position. 
Consequently, we start our SC generation algorithm by interpolating SC keypoints among the same HL (parent joints) and then continue to the next HL (child joints). Our SC Keypoint Generation algorithm is illustrated in Algorithm \ref{alg:SC-gen}. 
Two examples of our SC keypoint generation outputs are in Fig. \ref{Fig:classroom} (right) and Fig. \ref{Fig:features} (right). 

The skeleton pose and facial landmark extraction algorithms that we used, \cite{cao2019PAMI} and \cite{bulat2017ICCV} respectively, can relatively interpolate the missing parts when partial occlusion happens. To avoid the missing regions in the human body further, we only used the upper body of the human with HL $O$=4 (as described in Section \ref{Sec:ID}). We observed that the upper body of the human provides sufficient skeleton information to model different age groups.
Additionally, our SC keypoint generation algorithm can interpolate the missing keypoints of a side of the human body given the opposite side. For example, it can interpolate the right arm joint given the available left arm joint. In the extreme and less common cases when the facial landmark and skeleton pose extraction algorithms fail and also the upper body part also is not available, we set the missing values to zeros.

\vspace{10pt}

\begin{algorithm}
\setstretch{0.7}
\caption{SC Keypoint Generation}\label{alg:SC-gen}
\begin{algorithmic}
\Require The detected skeleton joints, $D$ 
\Ensure SC keypoints, $Z$ 
\State $i = 0 $ 
\State Add $D$ to $Z$ \Comment{all the detected joints are added to $Z$}
\While{$i \leq O$}  \Comment{$O$ is the number of HL in the skeleton}
\State $j = 0 $ 
\While{$j \leq J$}  \Comment{$J$ is in the same HL}
\If{$d^j(o_i)$ exists}
\If{$d^{j+1}(o_i)$ exists} \Comment{$d^{j}$ and $d^{j+1}$ are neighboring nodes}
    \State Add Interpolation($d^j(o_i), d^{j+1}(o_i)$) to $Z$ \Comment{in the same HL}
\EndIf
\EndIf
\State $j \gets j+1$  

\If{$d^{j}(o_{i+1})$ exists} \Comment{check neighboring nodes from next HL}
    \State Add Interpolation($d^j(o_i), d^{j}(o_{i+1})$) to $Z$ \Comment{in two neighboring HL}
\ElsIf{$D^{j}(o_{i+1})$ does not exist}
\If{$D^{j}(o_{i+1})$ is not end-effector} \Comment{like wrists and feet}
\State Add $0$ to $Z$  
\EndIf
\EndIf
\EndWhile
\State $i \gets i+1$  
\EndWhile
\end{algorithmic}
\end{algorithm}

\subsection{Facial Skeletal-Cosmetic Graph Construction}
\label{Sec:SCGC}
After obtaining $Z_i$ (SC keypoints) and $F_i$ (facial keypoints) in the previous steps, we construct a graph for each image sample $I_i$ to provide input to our TAA-GCN. We create the initial sets of feature vectors for $F_i$ and $Z_i$ as $X^F_i$ and $X^Z_i$, respectively. To create the sets of feature vectors, $\forall z_k \in Z_i$ and $\forall f_k \in F_i$, we assign an initial feature vector of $x^f_k \in X^F_i$ and $x^z_k \in X^Z_i$, respectively. Specifically, $x^f_k$ and $x^z_k$ are created by concatenating the patch of pixels around the keypoints $f_k$ and $z_k$ and their 2D coordinates. We tile the 2D coordinates of keypoints to match the patch of pixels size. In this hybrid feature representation, the patches of pixels provide cosmetic (clothing) information, while 2D coordinates give information about skeletal structures and postures.   

\begin{figure}[!htbp]
	\centering
		\includegraphics[height=0.2\textheight]{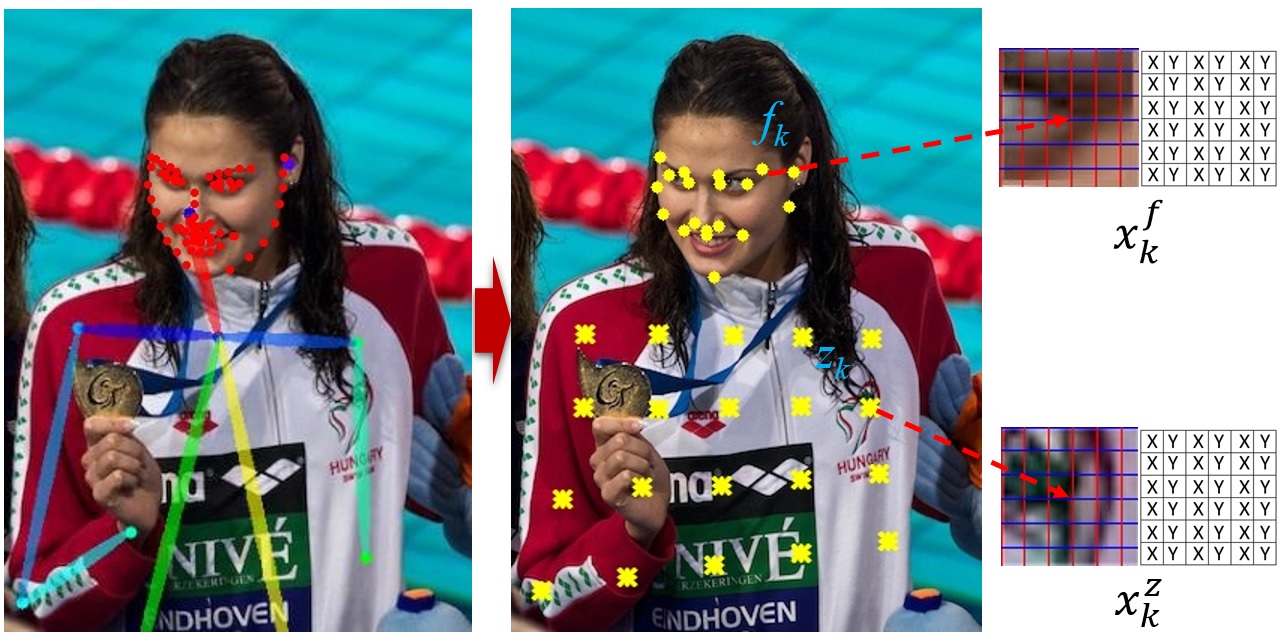}
	\caption{The left image shows the detected skeleton joints and facial keypoints; the right image indicates the final keypoints after using facial keypoint selection and SC keypoint generation algorithms. For each facial and SC keypoint $f_k$ and $z_k$, we assign an initial feature vector $x^f_k$ and $x^z_k$ by concatenating the patch of pixels (around each keypoint) and the 2D coordinate of the keypoint, ($X$, $Y$). In this feature representation, the 2D joint coordinates serve as the skeleton/posture structure, and the patches of pixels represent the clothing. ~\label{Fig:features}}
\end{figure}

An example of the aforementioned feature representation for each keypoint can be seen in Fig. \ref{Fig:features} (the image is collected from the UTKFace dataset). Each $x^f_k$ and $x^z_k$ then are converted to 1D vectors. Subsequently, the sets of feature vectors, $X^F_i$ and $X^Z_i$, are fed to the TAA-GCN. 
For two facial and SC graph nodes, we construct the initial adjacency matrices of $A^F$ and $A^Z$ by calculating the most correlated nodes. The correlation between each pair of nodes, $x_i$ and $x_j$, is calculated across all the dataset samples as: 

\begin{equation}
   \rho_{ij} = \dfrac {cov(x_i, x_j)}{\sigma x_i \cdot \sigma x_j }.
\end{equation}

So, $Aij = 1$ if $\rho_{ij}$ is among the top-$n$ correlated values, otherwise $Aij = 0$.  
Finally, we create the initial graph as $G_i=\{V_i; E_i\}$, where $V_i=\{X^F_i,X^Z_i\}$ and $E_i$ is obtained from $A = \{A^F,A^Z\}$. 

\vspace{10pt}

\subsection{Temporal-Aware Adaptive Graph Convolutional Network (TAA-GCN)}
\label{Sec:TAAGCN}
\begin{figure*}[!htbp]
	\centering 
		\includegraphics[width=1.0\textwidth]{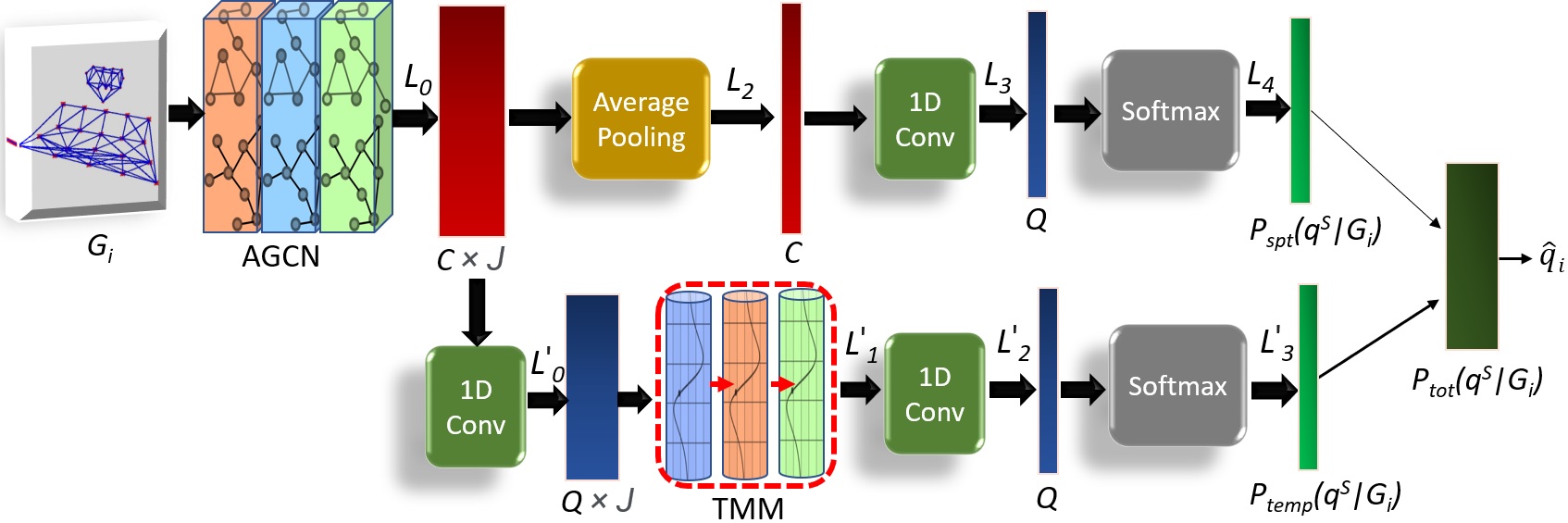} 
	\caption{Our proposed TAA-GCN pipeline which includes Adaptive GCN (AGCN) and a new Temporal Memory Module (TMM). Given the initial graph, $G_i$, the AGCN extract features, $L_0$ by refining the edges in $G_i$.  $L_0$ is then processed to compute the spatial age probability $P_{spt}$. Simultaneously, the TMM computes the non-ordinal temporal dependencies in $L_0$ to find the temporal age probability, $P_{temp}$. The final age prediction is obtained by a weighted summation of two spatial and temporal probabilities.  ~\label{Fig:TAA-GCN}}
\end{figure*}

The pipleline of our proposed TAA-GCN is shown in Fig. \ref{Fig:TAA-GCN}. The input of the TAA-GCN is a F-SC graph, $G_i=\{V_i; E_i\}$, and the output is an age label $\hat{q}_i \in q^S = \{q_t, t =0,1, \cdots , Q\}$, where $V_i = \{v_k \in \mathbb{R}^2, k=0, 1 \cdots J\}$, $E_i = \{E^F_i, E^Z_i\}$, and $Q$ is the number of age labels (maximum age). Here, $J= M' + N'$ is the total number of nodes, $E^F_i$ and $E^Z_i$ are facial and SC graph edges, respectively. 
Our TAA-GCN includes several Adaptive Graph Convolutional Layers (AGCL) to refine $E_i$ so that the updated graph edges, $\hat{E_i}$, become a more effective representation of connectivity among $V_i$. Specifically, by updating the graph edges, the TAA-GCN accommodates the variance in $E_i$, which is caused by variance in facial and SC graphs according to different faces, skeletons, postures, and clothing. 

The AGCN learns the spatial information in facial and SC graphs and outputs $L_0$, a $C \times J$ feature vector, where $C$ is the number of channels. $L_0$ is processed through Average Pooling (AP), 1D convolutional (Conv) and Softmax layers to output the spatial age probability $P_{spt}(q^S|G_i)$. The above spatial dataflow is also illustrated in Eq. \ref{EQ:prob_spt}, where $L_0 = AGCN(G_i)$.
The architecture of the AGCN which include several AGCLs is shown in Fig. \ref{Fig:AGCN}.  

\begin{figure}[!htbp]
	\centering 
		\includegraphics[height=0.12\textheight]{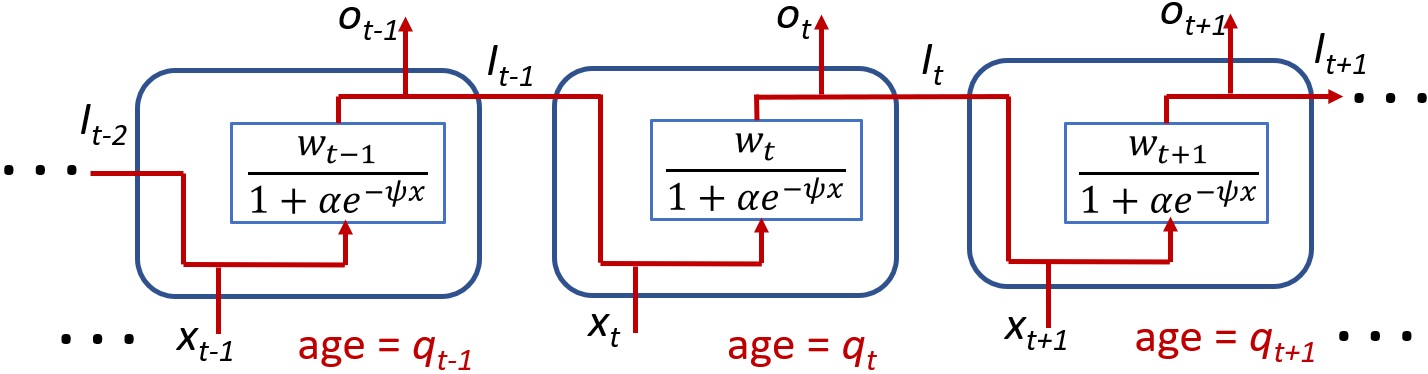} 
	\caption{Our proposed Temporal Memory Module. With using our new adaptive activation function including the trainable parameter $W_i$ and $\psi$, we extract non-ordinal temporal dependencies in age features, $x_t$, for various ages, $q_t$.  ~\label{Fig:TMM}}
\end{figure}

Simultaneously, the TMM processes a $Q \times J$ feature vector to learn the non-ordinal temporal dependencies among different ages. The TMM outputs the temporal features, $L'_1$, a $Q \times J$ feature vector. $L'_1$ then passes through convolutional and Softmax layers to compute $L'_3$, which is the temporal age probability $P_{temp}(q^S|G_i)$. The aforementioned temporal dataflow is also shown in Eq. \ref{EQ:prob_temp}. The architecture of the TMM can be seen in Fig. \ref{Fig:TMM}.

\vspace{10pt}

\begin{equation}
\label{EQ:prob_spt}
    P_{spt}(q^S|G_i) = Softmax(Conv(AP(L_0)))
\end{equation}

\begin{equation}
\label{EQ:prob_temp}
    P_{temp}(q^S|G_i) = Softmax(Conv(TMM(Conv(L_0))))
\end{equation}

Lastly, we compute the final age prediction as is shown in Eq. \ref{EQ:prob_tot}. 
The final loss $Loss_{F}$ can be seen in Eq. \ref{EQ:loss_tot}, where MSE is  Mean Square Error, $q_{g}$ is the ground truth value for age, and $\omega$ is an adjustment weight parameter. Additionally, $\hat{q}_{spt}$ and $\hat{q}_{temp}$ are spatially and temporally predicted ages, respectively.

\begin{equation}
\label{EQ:prob_tot}
    P_{tot}(q^S|G_i) = \omega \cdot P_{spt}(q^S|G_i) + (1 - \omega) \cdot P_{temp}(q^S|G_i)
\end{equation}

\begin{equation}
\label{EQ:loss_tot}
    Loss_{F} = \omega \cdot MSE(\hat{q}_{spt} - q_{g}) + (1 - \omega) \cdot MSE(\hat{q}_{temp} - q_{g})
\end{equation}

We will explain both TMM and AGCL in the following:

\vspace{10pt}

\begin{figure}[!htbp]
	\centering 
		\includegraphics[height=0.15\textheight]{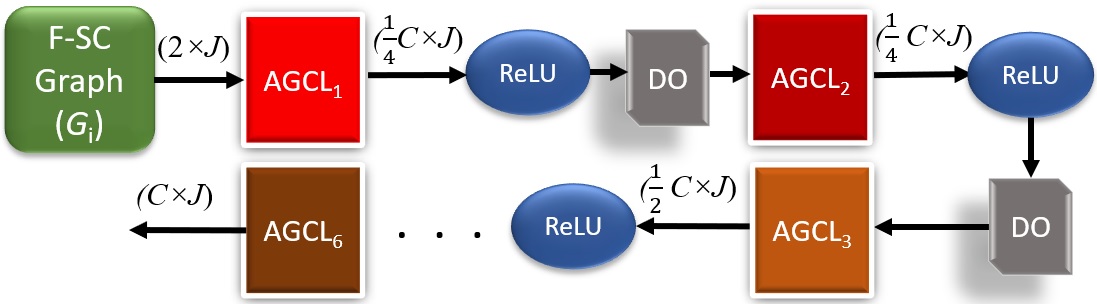} 
	\caption{The architecture of the proposed Adaptive Graph Convolutional Network (AGCN). Here, DO signifies a Dropout layer,  $C=256$, and the activation function is the Rectified Linear Unit (ReLU). ~\label{Fig:AGCN}}.
\end{figure}

\textbf{Temporal Memory Modules.}

Standard temporal networks follow an ordinal temporal structure to capture temporal dependencies in sequences. Such a temporal structure is irrelevant in age estimation since there is no explicit temporal connection between consecutive age samples. In contrast, our TMM can capture temporal dependencies among temporally non-ordinal age samples. 
As shown in Fig. \ref{Fig:TAA-GCN}, the input of the TMM is a $Q \times J$ feature vector, where each raw represents different ages as $\{q_t, t =0,1, \cdots , Q\}$. The TMM captures the temporal dependencies among different $q_t$ in a recurrent manner which is shown in Fig. \ref{Fig:TMM}. In this figure, $x$ is the input, $o$ is the output, and $I$ is the hidden state. 

\vspace{10pt}

We exploit the fact that people of similar ages share more similarities than different ages. For example, people aged 20 and 21 look more similar, wear more similar clothing and have more similar poses than those aged 20 and 60. Our proposed TMM can capture the dependencies between both close age groups such as 20 and 21 as well as distant age groups like 20 and 60.
By considering such temporal dependencies among age groups, we assist the age estimation network to (1) decrease the intra-class variations by finding similarities among the same ages, and (2) increase the inter-class variations by learning the differences among different ages. Especially, the latter is crucial to avoid significant errors such as confusion between ages 20 and 60 that lead to a remarkable decrease in the overall network performance. To make our TMM aware of such age group dependencies, we propose an adaptive activation function, $\phi$, illustrated in Eq. \ref{Eq:phi}.

\begin{equation}
\label{Eq:phi}
    \phi = \frac{W}{1+ \alpha e^{-\psi x}}, \; \; \; \; \;\psi = \frac{1}{|q_t-q_g|}
\end{equation}

In the above, $q_g$ is a ground truth age, $\alpha > 1$ is an adjustment parameter, and $W = \{w_t, t=0,1 \cdots, Q\}$ are adaptive weight parameters. $\psi$ weights different age groups and ensures that during the training, the ages closer to the ground truth are assigned higher weights than distant ages. In the extreme cases, when two ages are the same, $|q_t-q_g| \longrightarrow 0$, then $\psi \longrightarrow \infty$, and so $\phi \longrightarrow w_t$. On the other hand, when $|q_t-q_g| \longrightarrow Q$, then $\psi \longrightarrow 0$, and so $\phi \longrightarrow \frac{w_t}{1 + \alpha}$. The adaptive weight parameters memorize this weighting procedure for different $q_t$. 
In the testing phase, $\psi$ is set to 1, and we use $W$ to weight different $q_t$.


\textbf{Adaptive Graph Convolutional Layer (AGCL).}

The graph convolutional operator is applied over each $v_k$'s adjacent nodes $\beta (v_k)$. The set of graph edges, $E_i$, includes all $\beta (v_k), v_k \in V_i$. Hence, finding appropriate $E_i$ is crucial for applying effective convolutions over $\beta (v_k)$ that maximizes the network performance. Due to the variance in facial and SC graphs for different faces, skeletons, postures, and clothing, finding appropriate $E_i$ is challenging. To solve this challenge, some researchers have suggested using dynamic temporal graphs for temporal problems such as skeleton-based action recognition \cite{korban2020ECCV}. Our case, however, is different because: (1) age estimation is a spatial problem, where many node candidates are equally important. For example, in a temporal sequence such as ``running"  few joints are involved in the action. In different ages, however many facial and SC keypoints have significant equal roles; (2) in contrast to the skeletal graphs \cite{korban2020ECCV}, our F-SC graph consists of two separate facial and SC graphs, with no explicit connection between them. 
So, instead of eliminating the nodes, we propose a weighting strategy that prioritizes the importance of node connectivity in $E_i$. Moreover, we utilize two separate weighting functions to refine the edges in facial and SC graphs.

Our AGCN consists of several AGCLs that refine $E_i$ separately for facial and SC graphs via updating the weight of each edge $e_{mn} \in E_i$ which connects $(v_m, v_n) \in V_i$. We define the relationship between the input ($f_{in}$) and the output ($f_{out}$) of each AGCL for each $v_m \in V_i$ as:

\begin{equation}
\label{EQ:GCN}
    f_{out}(v_m) = \sum_{v_n \in \beta(v_m)} f_{in}(v_n) \cdot (f_c(e_{mn}))
\end{equation}

In the above, $f_c$ is our proposed correlation weighting function and $e_{mn} = \{e^F_{mn}, e^Z_{mn}\}$, where $F$ and $Z$ represent for facial and SC graphs, respectively.  
The correlation weighting function, $f_c$, ensures that highly correlated nodes have larger weights on their connecting edges. To find the correlations among nodes we compute the cosine similarities ($C_{S}$) between the pair of nodes $v_m$ and $v_n$ as: 

\begin{equation}
    C_{S(v_m, v_n)} = \frac{v_m \cdot v_n}{	\lVert  v_m  \rVert \lVert  v_n  \rVert}
\end{equation}

We selected the cosine distance since compared to Euclidean distance is not sensitive to node vectors' magnitudes. Specifically, the Euclidean distance fails to capture the correlation between highly correlated nodes that are located distantly in the Euclidean space. For a similar reason, we could not use the city block distance.

To include the correlation information between each nodes $(v_m, v_n) \in V_i$, we update each edge $e_{mn} \in E_i$ using $f_{c}$. For SC and facial graphs, the facial and SC correlation weighting functions, $f_{c}^{Z}$ and  $f_{c}^{F}$, can be seen in Eq. \ref{EQ:ZCor} and  Eq. \ref{EQ:FCor}, respectively.

\begin{equation}
\label{EQ:ZCor}
    f_{c}^{Z}(e^Z_{mn}) = {W^Z} \cdot C_{S(v_m, v_n)} \cdot e^Z_{mn}
\end{equation}

\begin{equation}
\label{EQ:FCor}
    f_{c}^F(e^F_{mn}) = \frac{W^F}{1+ {C_{S(v_m, v_n)} \cdot e^F_{mn}}^{-\gamma_n}}
\end{equation}

In the above equations, $W^Z$ and  $W^F$  are trainable parameters. Since facial graphs might not change dramatically similar to SC graphs, for facial graphs, we magnify the correlation change in $e^F_{mn}$ by a logistic function with the power of $\gamma_n$. 
Specifically, $\gamma_n \in 1 \leq \mathbb{Z} \leq K$ is the order of $e^F_{mn} \in E_i(\beta(v_m))$, where $K$ is the number adjacent nodes in $\beta(v_m)$. Namely, for each  $v_n \in \beta(v_m)$, $e^F_{mn}$ is ordered based on the correlation value between $v_n$ and $v_m$. 
So, the optimal values of $\gamma_n$ for  $v_m$ and its neighboring nodes $v_n$ is calculated based on order of correlation values between neighboring nodes. We used this optimal weighting because it is consistent for all nodes $V_i$ since $K$ is a fixed parameter.

We separately update $e^F_{mn}$ and $e^Z_{mn}$ by defining two separate correlation weighting functions ${f_c}^{F}$ and ${f_c}^Z$ to ensure that the original connectivity importance in the facial and SC graphs are preserved. 
Each $e_{mn}$ is adaptively updated after each convolutional operation between $\beta(v_m)*v_{m} \cdot f_c(e_{mn})$ which also updates $W^Z$ and $W^F$ in each gradient descent iteration.

\subsection{Implementation Details}
\label{Sec:ID}
Table \ref{Tab:impl} summarizes the implementation details of our proposed pipeline referenced in relevant sections of this paper. 

\vspace{10pt}

\begin{table}[h!tbp]
	\centering
	\caption{Implementation details of our proposed pipeline with associated paper section references.} \label{Tab:impl} 
\resizebox{\textwidth}{!}{\begin{tabular}{ccc}
\hline
Parameter & Value & Section \\ \hline
Number of Initial Facial Keypoints ($N$) & 68 & \ref{Sec:overview} \\
Number of Skeleton Joints ($M$) & 18 & \ref{Sec:overview} \\
Number of Selected Facial Keypoints ($N'$) & 19 & \ref{Sec:FKS} \\
Number of Neighboring Facial Keypoints  ($K $) & 5 & \ref{Sec:FKS}\\
Facial Keypoints Selection Weighting Parameter ($\eta$) & {0.4} & \ref{Sec:FKS} \\
Number of SC Keypoints ($M'$) & 20 & \ref{Sec:SCKG} \\
Number of Hierarchical Levels ($O$) & 4 & \ref{Sec:SCKG} \\ 
Size of Patch of Pixels & 32x32 pixels & \ref{Sec:SCGC} \\
Number of AGCLs & 6 &  \ref{Sec:TAAGCN} \\
Number of Channels in AGCL (each layer)  & (64, 64, 128, 128, 256, 256) &  \ref{Sec:TAAGCN} \\
Number of Layers in TMM & 5 &  \ref{Sec:TAAGCN} \\
Number of Channels in TMM (for all layers) & 64 &  \ref{Sec:TAAGCN} \\
Learning Rate & $1e^{-5}$ &  \ref{Sec:TAAGCN} \\
Number of Training Epochs & 300 &  \ref{Sec:TAAGCN} \\
Optimizer & ADAM & \ref{Sec:TAAGCN} \\
Weight Decay & $1e^{-6}$ & \ref{Sec:TAAGCN}\\
Loss Adjustment Weight ($\omega$) & 0.65 & \ref{Sec:TAAGCN} \\
\hline
\end{tabular}}
\end{table}

\section{Experimental Results}

\subsection{Datasets}
For our major comparative experiments, we used four age datasets, CACD \cite{chen2015PRA}, MORPHII \cite{chen2015PRA}, UTKFace \cite{zhifei2017cvpr}, and FG-NET \cite{ranking2014ECCV}, in our experiments. For the FG-NET, MORPHII, and CACD datasets we only used our facial graphs since they do not include the human body. 

\textbf{UTKFace.} 
Consists of 24,000 images of 116 ages. We used 20,000 samples for training and 4,000 samples for testing. For now, it is the only major publicly-available age dataset that includes uncropped images with sufficient views of clothing. So, the dataset is well matched with our network when both facial and SC graphs are used. 

\textbf{MORPHII.} 
 Includes 55,000 samples of 13,000 subjects ranging from age 16 to 77. We used 44,000 samples for training and 11,000 samples for testing.

\textbf{CACD.} 
Includes 160,000 sample images of 2,000 celebrities from 49 age categories. In our experiments, the training, testing, and validation sizes are 127,000, 31,000 and 12,000 samples, respectively. As the CACD dataset only includes faces, we tested this dataset only on our facial graphs.

\textbf{FG-NET.}
Consists of about 1,000 samples from 69 different ages, and 82 individuals. We used 900 samples for training and 100 samples for testing. This dataset also includes only face images. 

\subsection{Comparative Results}
We compared our method with the state-of-the-art approaches on four datasets, UTKFace, CACD, MORPHII, and FG-NET, with the results shown in Table \ref{Tab:compareUTK} and Table \ref{Tab:compareFG}, respectively. 
We used the Mean Absolute Error (MAE), an standard evaluation metric in age estimation. The description of MAE is shown in Equation \ref{eq:mae}, where $y_i$ is the estimated age for each test sample $i$,  $y_i^G$ is the ground truth, and $N$ is the number of test samples. 

\begin{equation}
\label{eq:mae}
    MAE = \sum_{i=0}^{N}{|y_i - y_i^G|}
\end{equation}

Our proposed approach outperformed the state-of-the-art methods on all four benchmarks. Specifically, for the UTK-Face dataset, with an MAE of 3.48, our TAA + F-SC graph outperformed the best previous record by approximately one year in age accuracy. For the MORPHII dataset, we obtained an MAE of 1.69, which outperformed the other methods. For the CACD dataset, which is the largest existing age dataset, we obtained an MAE of 4.09, which improves upon the currently best results by 0.54 years in age accuracy. 
Our proposed method outperforms the existing approaches that did not use the IMDB-WIKI dataset pre-trained weights/features on the FG-NET dataset.  We improved the current benchmark by 0.34 years. Among the methods that used IMDB-WIKI dataset for pre-training however, DAG-CNN \cite{taheri2019NC} achieved the best performance.
We outperformed the other works on two large MORPHII and CACD datasets, though many of them used IMDB-WIKI pre-trained weights/features.

\begin{table}[h!tbp]
	\centering
	\caption{Comparison of our method, TAA-GCN + Facial graph and the state-of-the-art methods on the MORPHII dataset.} \label{Tab:compareCACD} 
\begin{tabular}{ccc}
\hline
Team & Method & MAE \\ \hline
Lu et al. \cite{niu2016CVPR} & ORMO &  3.27 \\
Liu et al. \cite{liu2019TIP} & SAF &  2.97 \\
Zhang et al. \cite{zhang2019CVPR} & C3AE & 2.75 \\
Cao et al. \cite{cao2020PR} & RCOR &  2.64 \\
Yang et al. \cite{yang2018IJCAI} & SSR-Net &  2.52  \\
Pan et al. \cite{pan2018CVPR} & MVL & 2.51  \\
Shen et al. \cite{shen2017NIPS} & LDIR & 2.24 \\
Shen et al. \cite{shen2018CVPR} & DRF & 2.17 \\
\textbf{Ours} & \textbf{TAA-GCN+F} &\textbf{1.69} \\ \hline
\end{tabular}
\end{table}

\begin{table}[h!tbp]
	\centering
	\caption{Comparison of our proposed method, TAA-GCN + F-SC graph and the state-of-the-art methods on the UTKFace dataset.} \label{Tab:compareUTK} 
\begin{tabular}{ccc}
\hline
Team & Method & MAE \\ \hline
Rothe et al. \cite{rothe2018IJCV} & DEX & 4.31 \\
Yoshimura et al. \cite{yoshimura2020arx} & FOSS & 4.49 \\
Cao et al. \cite{cao2020PR} & RCOR & 5.39 \\
Al et al. \cite{al2020JJCIT} & SDTL & 4.86 \\
Li et al. \cite{li2020AAAI} & SAM & 4.77 \\
Berg et al. \cite{berg2021ICPR}  & DOR & 4.55 \\
Sun et al. \cite{sun2021info} & DCD & 4.47 \\
\textbf{Ours} & \textbf{TAA-GCN+F-SC} &\textbf{3.48} \\ \hline

\end{tabular}
\end{table}

\begin{table}[h!tbp]
	\centering
	\caption{Comparison of our method, TAA-GCN + Facial graph and the state-of-the-art methods on the CACD dataset.} \label{Tab:compareCACD} 
\begin{tabular}{ccc}
\hline
Team & Method & MAE \\ \hline
Lu et al. \cite{niu2016CVPR} & ORMO &  5.36 \\  
Cao et al. \cite{cao2020PR} & RCDR &  5.24 \\
Yang et al. \cite{yang2018IJCAI} & SSR-Net &  4.96 \\
Zhang et al. \cite{zhang2019CVPR} & C3AE & 4.88 \\
Rothe et al. \cite{rothe2018IJCV} & DEX &  4.78 \\
Shen et al. \cite{shen2017NIPS} & LDIR &  4.73  \\
Shen et al. \cite{shen2018CVPR} & DRF & 4.63  \\
\textbf{Ours} & \textbf{TAA-GCN+F} &\textbf{4.09} \\ \hline
\end{tabular}
\end{table}

\begin{table}[h!tbp]
	\centering
	\caption{Comparison of our method and the state-of-the-art methods on the FG-NET dataset. The results for this small dataset are reported for two categories: (1) methods that employed training from scratch rather than pre-trained weights and features; methods that used IMDB-WIKI pre-trained weights/features.} \label{Tab:compareFG} 
\resizebox{\textwidth}{!}{\begin{tabular}{cccccc}
\hline
Methods NOT pre-trained &  DE \cite{han2014PAMI} & CS-LBF \cite{lu2015TIP} &
GADF \cite{liu2017PR} & SAF \cite{liu2019TIP} & \textbf{Ours} \\ \hline
MAE & 4.80 & 4.36 & 3.93 & 3.92 & \textbf{3.58} \\ \hline
Methods pre-trained & DEX  \cite{rothe2018IJCV} & MVL \cite{pan2018CVPR} &
C3AE  \cite{zhang2019CVPR} & DAG-CNN \cite{taheri2019NC} &  \\ \hline
MAE & 4.63 & 4.10 & 4.09 & \textbf{3.05} & \\ \hline
\end{tabular}}
\end{table}

\subsection{Ablation Study}
We conducted an ablation study to evaluate the impact of the constituent components of our proposed pipeline on the overall age estimation performance. For our study, we define some abbreviations that are shown in Table \ref{Tab:terms}. We carried out our ablation study on the UTKFace dataset because it includes uncropped images with visible clothing, allowing us to test various types of graphs.
We compare two facial and SC graphs (keypoints and patches of pixels) separately in Table \ref{Tab:abl1}. The overall age estimation performance changed when different graphs (facial or skeletal-cosmetic) were used.

We encode the skeletal and posture information with our Skeletal-Cosmetic Keypoints (SCK), while we obtain the clothing information from our Skeletal-Cosmetic Image Patches (SCIP). As can be seen in Table \ref{Tab:abl1}, using the skeleton/posture (SCK) information alone led to slightly better performance (MAE=5.50) compared to using clothing information alone (MAE=5.56). The combination of skeleton/posture and clothing information (SCK+CSIP), however, yields the best performance (MAE=5.45).
Table \ref{Tab:abl2} illustrates the impact of the combination of facial and SC graphs (skeleton/posture and clothing information) on the overall age estimation performance.  Overall, combining all facial, skeleton/posture, and clothing features (FK+SCK+FPP+SCIP) led to superior performance.

Without facial information, and by using only SC graphs and SCK+SCIP features, still our proposed approach achieved competitive performance (with an MAE of 5.45 according to Table \ref{Tab:abl1}). Our age estimation algorithm works well without facial information, using only skeleton, posture and clothing information. So, our proposed algorithm is practical in real-world scenarios such as surveillance systems, where facial information is partially available.  

 We also analyzed the impact of the constituent components of our proposed network, which is shown in Table \ref{Tab:abl3} (with facial keypoint selection) and Table \ref{Tab:abl4} (without facial keypoint selection). We achieved the best performance (with an MAE of 3.48) when we jointly used all our proposed modules (TAA-GCN) with our facial keypoint selection algorithm. Moreover, we evaluated the impact of the trainable facial and SC graph edge parameters, $W^F$ and $W^Z$, on the overall age estimation performance illustrated in Table \ref{Tab:abl5}. To evaluate the generalization of our proposed approach, we conducted a cross-dataset evaluation on the MORPHII and CACD datasets which have sufficient numbers of samples for appropriate training. The results are shown in Table \ref{Tab:abl7}.

\begin{table*}[h!tbp]
	\centering
	\caption{The abbreviations used in the ablation study.} \label{Tab:terms} 
\begin{tabular}{cc}
\hline
Abbreviation & Definition \\ \hline
FK & Facial Keypoints (2D coordinates)\\
FPP & Facial Patches of Pixels \\
SCK & Skeletal-Cosmetic Keypoints (2D coordinates)\\
SCIP & Skeletal-Cosmetic Image Patches \\
TAA-GCN & Temporal-Aware Adaptive GCN \\
AGCN & Adaptive GCN (no temporal awareness) \\
TA-GCN & Temporal-Aware GCN (no adaptivity) \\
GCN  & Baseline GCN (no TA and adaptivity) \\
\hline
\end{tabular}
\end{table*}

\begin{table*}[h!tbp]
	\centering
	\caption{The impact of facial/SC graphs and their keypoints/patches of pixels, individually, on the overall age estimation performance. The skeletal and posture information are encoded using the Skeletal-Cosmetic Keypoints (SCK). The clothing information are encoded using Skeletal-Cosmetic Image Patches (SCIP).} \label{Tab:abl1} 
\resizebox{\textwidth}{!}{\begin{tabular}{ccccccc}
\hline
Modules &
FK &
FPP &
FK+FPP &
SCIP &
SCK &
SCK+SCIP \\
 \hline
 Modality &
face &
face &
face &
clothing &
skeleton/posture &
clothing+skeleton/posture \\
\hline
MAE &
5.51 &
5.05 &
4.18 &
5.56 &
5.50 &
5.45 \\
 \hline
\end{tabular}}
\end{table*}

\begin{table*}[h!tbp]
	\centering
	\caption{The impact of the skeletal and posture (SCK) and clothing (SCIP) information, combined with facial information on the overall age estimation performance.} \label{Tab:abl2} 
\resizebox{\textwidth}{!}{\begin{tabular}{cccc}
\hline
Modules &
FK+FPP+SCK &
FK+FPP+SCIP &
FK+FPP+SCK+SCIP  \\
 \hline
Modality &
face+skeleton/posture &
face+clothing &
face+skeleton/posture+clothing\\
 \hline
MAE &
3.64 &
3.75 &
3.46 \\
 \hline
\end{tabular}}
\end{table*}

\begin{table}[h!tbp]
	\centering
	\caption{The impact of the constituent components of our proposed network on the overall age estimation performance when the facial keypoint selection algorithm \textbf{is} used.} \label{Tab:abl3} 
\begin{tabular}{ccccc}
\hline
Modules & TAA-GCN & AGCN & TA-GCN & GCN \\ \hline
MAE & 3.48 & 4.83 & 4.02 & 5.33 \\
\hline
\end{tabular}
\end{table}

\begin{table}[h!tbp]
	\centering
	\caption{The impact of the constituent components of our proposed network on the overall age estimation performance, when the facial keypoint selection algorithm \textbf{is not} used.} \label{Tab:abl4} 
\begin{tabular}{ccccc}
\hline
Modules & TAA-GCN & AGCN & TA-GCN & GCN \\ \hline
MAE & 4.01 & 5.60 & 4.78 & 5.98 \\
\hline
\end{tabular}
\end{table}

\begin{table}[h!tbp]
	\centering
	\caption{The impact of the trainable facial and SC graph edge parameters, $W^F$ and $W^Z$, on the overall age estimation performance.} \label{Tab:abl5} 
\resizebox{\textwidth}{!}{\begin{tabular}{ccccc}
\hline
trainable parameter $W^F$ & adaptive & adaptive &  non-adaptive & non-adaptive \\ \hline
trainable parameter $W^Z$ & adaptive & non-adaptive & adaptive & non-adaptive \\ \hline
MAE & 3.48 & 4.43 & 4.20 & 4.78 \\
\hline
\end{tabular}}
\end{table}

\begin{table}[h!tbp]
	\centering
	\caption{Cross-dataset analysis between the MORPHII and CACD datasets.} \label{Tab:abl7} 
\begin{tabular}{ccc}
\hline
Scenario & Trained on MORPHII & Trained on CACD  \\ 
& tested on CACD & tested on MORPHII \\ \hline
MAE & 5.56 & 3.02 \\
\hline
\end{tabular}
\end{table}

\vspace{30pt}

\subsection{Images in-the-Wild}

\begin{figure}[!htbp]
	\centering
		\includegraphics[height=0.17\textheight]{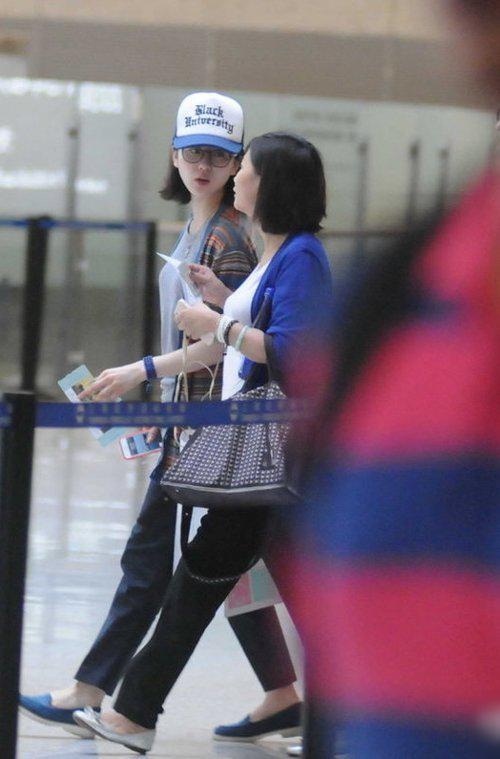}
		\includegraphics[height=0.17\textheight]{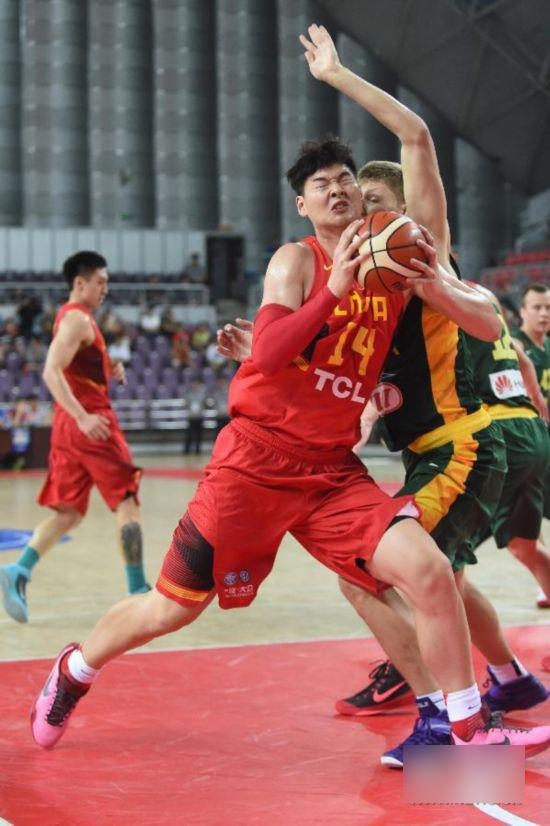}
		\includegraphics[height=0.17\textheight]{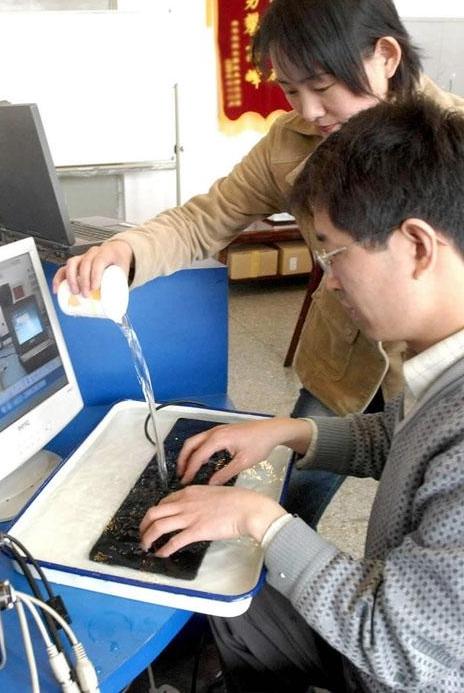}
             \includegraphics[height=0.17\textheight]{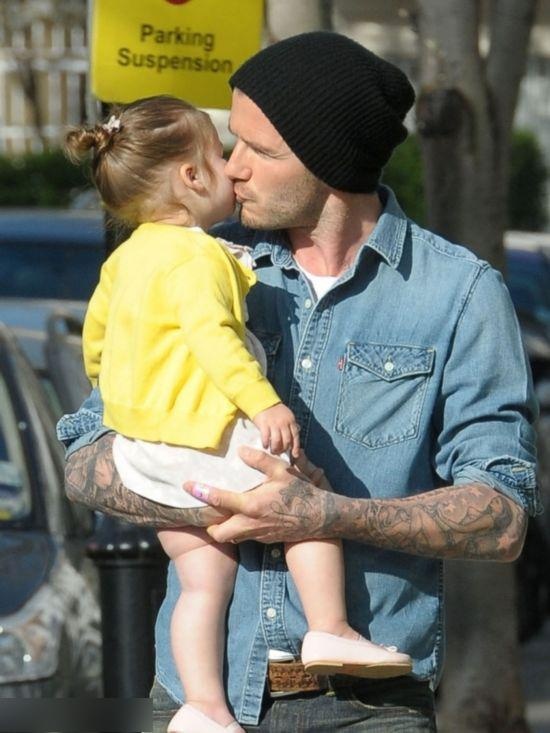}
	\caption{Some image samples form the Relative Human dataset with a variety of partial facial views captured in-the-wild that also resembles surveillance conditions.} ~\label{Fig:RH}
\end{figure}

The images from the UTKFace \cite{zhifei2017cvpr} that we used in our experiments are captured in-the-wild. To further evaluate our proposed approach in-the-wild condition, we also used the Relative Human dataset \cite{sun2022CVPR}  which includes a variety of images captured in-the-wild. The Relative Human dataset includes 24800 image samples with a variety of partial face views commonly due to the impact of different camera angles or subjects/objects overlap in-the-wild. Such the in-the-wild condition also shares many similarities with surveillance conditions, where people are captured from different camera viewpoints. Some image examples are shown in Fig. \ref{Fig:RH}. The dataset also includes both human face and body which is useful in our experiments. The Relative Human dataset provided the label set consisting of four age groups ``baby'', ``kid'', teenager'', and ``adult''. 
For evaluation, we used the ``top-1 score'', a common metric in classification problems, that  matches the top class with the highest probability and the target label. We reported the age estimation results for different combinations of facial and SC graphs (face, skeleton/posture and clothing information) in Table \ref{Tab:RH1}. We also compared our method to two other approaches, DEX \cite{rothe2018IJCV} and SSR \cite{yang2018IJCAI} in Table \ref{Tab:RH2}. As can be seen, our approach with the combination of all features (FK+FPP+SCIP+SCK) achieved a top-1 score of 97.80\% which is significantly  higher than those scores for other methods.

\begin{table*}[h!tbp]
	\centering
	\caption{The impact of facial/SC graphs and their keypoints/patches of pixels on the age estimation performance on the Relative Human dataset.} \label{Tab:RH1} 
\resizebox{\textwidth}{!}{\begin{tabular}{ccccc}
\hline
Modules &
FK+FPP &
SCIP &
SCK &
SCK+SCIP \\
 \hline
 Modality &
face &
clothing &
skeleton/posture &
clothing+skeleton/posture \\
\hline
Top-1 score &
94.35\%  &
91.85\%  &
91.10\%  &
91.98\% \\
 \hline
\end{tabular}}
\end{table*}

\begin{table*}[h!tbp]
	\centering
	\caption{Comparison of our proposed method to other approaches on the Relative Human dataset.} \label{Tab:RH2} 
\begin{tabular}{cccc}
\hline
Method &
\textbf{Ours (FK+FPP+SCIP+SCK)} &
DEX \cite{rothe2018IJCV} &
SSR \cite{yang2018IJCAI} \\
\hline
Top-1 score &
\textbf{97.80\%} &
90.03\% &
92.31\% \\
 \hline
\end{tabular}
\end{table*}

\subsection{Blurring Effects and Facial Expressions}

\begin{figure}[!htbp]
	\centering
	\begin{tabular}{c c c}
		\includegraphics[height=0.1\textheight]{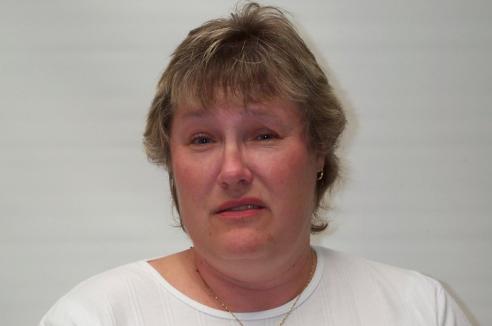} & \hspace{-13pt}
		\includegraphics[height=0.1\textheight]{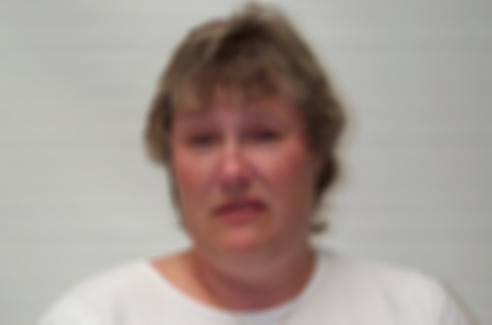} & \hspace{-13pt}
		\includegraphics[height=0.1\textheight]{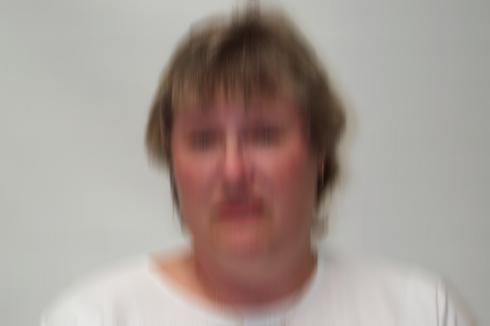} \\
  		\includegraphics[height=0.1\textheight]{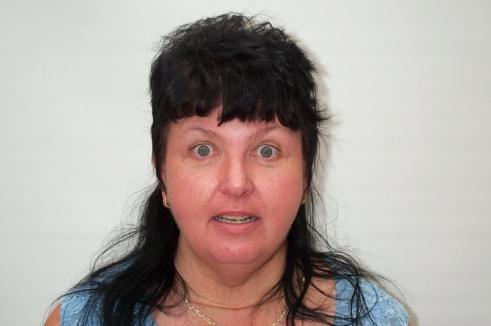} & \hspace{-15pt}
		\includegraphics[height=0.1\textheight]{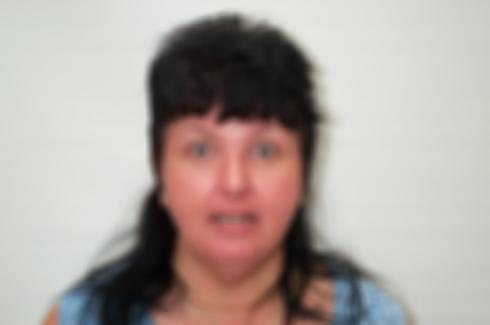} & \hspace{-15pt}
		\includegraphics[height=0.1\textheight]{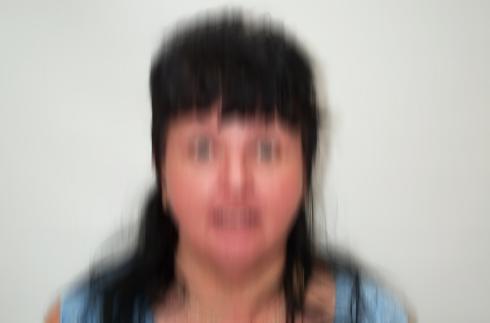} \\
  		\includegraphics[height=0.114\textheight]{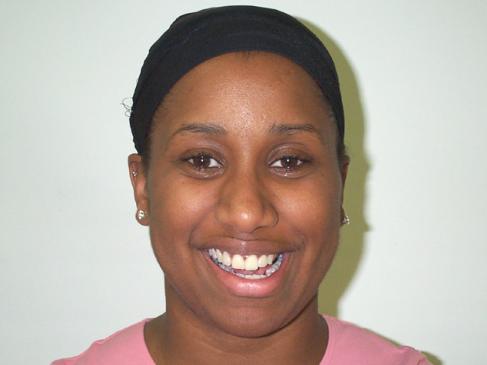} & \hspace{-15pt}
		\includegraphics[height=0.114\textheight]{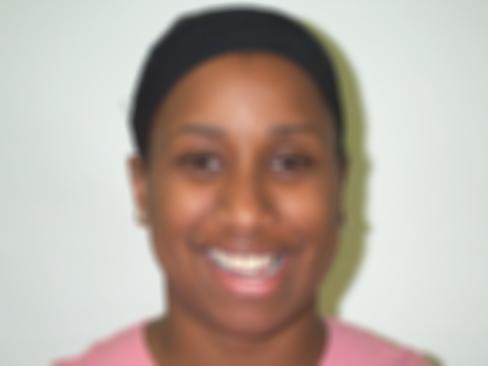} & \hspace{-15pt}
		\includegraphics[height=0.114\textheight]{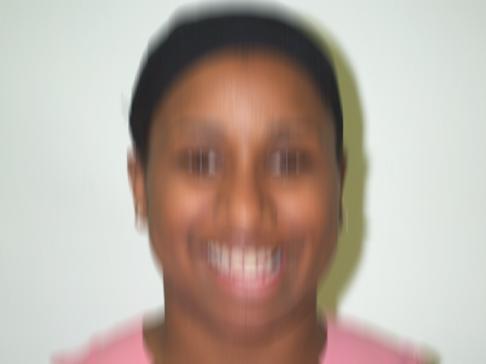} \\
            (a) & (b) & (c) \\
   \end{tabular}
	\caption{Some image samples form the PAL dataset with two blurring effects under different facial expressions. (a): Original image, (b): added Gaussian blurring effect, and (c): added motion blurring effect. The facial expressions are: (up): sad, (middle): surprise, and (down): happy. } ~\label{Fig:PAL}

\end{figure}

To simulate the reduced-quality image capture in-the-wild or under surveillance conditions, following \cite{nguyen2015Sym} and \cite{kang2018Sym}, we evaluated our proposed method under several blurring effects. We tested our algorithm on the PAL dataset \cite{minear2004PAL} which includes 3000 image samples with a variety of facial expressions. We synthetically added Gaussian and motion blurring effects to the image samples of the dataset to emulate those effects in real-life scenarios. Some image examples from the PAL dataset with three facial expressions, ``sad'', ``surprise'', and ''happy'' and synthetically added blurring effects are shown in Fig. \ref{Fig:PAL}. We compared our method with two other strategies and showed the results in Table \ref{Tab:PAL2}. As can be seen, our method remarkably outperforms the other approaches which shows the reliability of our proposed approach under blurring effects which is a common condition in-the-wild or in surveillance scenarios.

\begin{table*}[h!tbp]
	\centering
	\caption{Comparison of our proposed method to other approaches on the PAL dataset under different blurring effects that simulates those effects in-the-wild or under surveillance conditions.} \label{Tab:PAL2} 
\begin{tabular}{ccc}
\hline
Method & MAE (Gaussian Blur) & MAE (Motion Blur) \\
\hline
\cite{nguyen2015Sym} &
6.42 &
6.48 \\
\hline
\cite{kang2018Sym} &
6.0 &
6.0 \\
\hline
\textbf{Ours} &
\textbf{3.49} &
\textbf{3.70} \\
 \hline
\end{tabular}
\end{table*}

\section{Conclusions}

We proposed a new graph representation for age estimation using skeleton joints and facial keypoints. This new representation yields more relevant information than raw images and is more reliable under different viewpoints and facial expressions. We also suggested a new Temporally-Aware Adaptive Graph Convolutional Network with two improvements (1) it captures non-ordinal temporal dependencies in different ages which is not possible with standard temporal networks, and (2) it adaptively refines facial and Skeletal-cosmetic graphs edges to accommodate the variance in appearance. Furthermore, we proposed to use skeleton structure, posture, and clothing information in the age estimation solution. This rich set of features accommodates significant performance improvements when the face is only partially visible in real-life scenarios.
Our method outperformed the state-of-the-art approaches on four public benchmarks, including the UTKFace dataset, whose images are captured in-the-wild.
We also further tested the reliability of our proposed age estimation algorithm in uncontrolled environments in two more scenarios: (1) the images captured in-the-wild from the Relative Human dataset and (2) the synthetically blurred images from the PAL dataset under a variety of facial expressions.

Our new graph representation of soft biometrics, including the skeleton, posture, clothing, and face, can be used as a backbone by other researchers in the field. Such a graph structure is a powerful representation of the skeleton and face in tandem, since graph nodes and edges can effectively describe facial landmarks, and skeleton joints and encode their connectivity information. We also are the first to introduce a complete experimental setup for age estimation in-the-wild. Such an experimental setup can be used as a standard benchmark in the future. Moreover, our new Temporal Memory Module (TMM) can be exploited in any research problem, such as age estimation that involves computing non-ordinal temporal dependencies.

\subsection{Limitations}
Here is our work's main limitations:
\begin{itemize}
    \item \textit{Severe occlusion:} While our method is effective when the human face and body are partially occluded, it can fail when \textit{both} the face and body are severely occluded. However, such a severe occlusion is unlikely.
    \item \textit{Non-conventional facial expressions:} Our facial keypoint selection algorithm can accommodate a variety of facial expressions. Nevertheless, our proposed algorithm might be less effective in handling non-conventional facial expressions.   
    \item \textit{Complex human poses:} Although our Adaptive Graph Convolutional Layer (AGCL) can accommodate diverse standard and non-standard human poses, it might be less effective in handling extremely complex human poses in sports activities such as gymnastics or martial arts. It is because, in many of these activities, the skeleton configuration of human changes significantly. So, the neighboring skeleton joints can highly vary in such sport activities. 
    \end{itemize}

\subsection{Future Work}
We suggest some future work to solve the limitations explained above:
\begin{itemize}
    \item \textit{Modeling human sub-parts:} Although our graph-based model can model several human face and body sub-parts, such as eyes and noise, the network's loss function does not depend independently on human sub-parts. We suggest modeling human sub-parts independently with separate loss functions to handle severe face and human body occlusion. Such independent human sub-parts modeling is more effective when only a sub-part of the human face or body is visible. This can be achieved, i.e., by designing multiple graph convolutional streams for several human sub-parts. 
    \item \textit{Adaptive facial keypoint selection:} Although our facial keypoint selection algorithm is data-driven, it is not based on a learning process. To accommodate a variety of cross-cultural and non-conventional facial expressions, we recommend a more adaptive model, preferably using a separate deep learning module to select more effective facial keypoints.   
    \item \textit{Adaptive skeletal-cosmetic keypoint generation:} Our skeletal-cosmetic keypoint generation algorithm selects fixed and spatially consistent keypoints based on human skeleton structure. We suggest an adaptive way to generate such keypoints to handle various complex human poses. 
    \end{itemize}

\textbf{Acknowledgements}

This material is based upon work supported by the National Science Foundation under Grant No. 2000487 and the Robertson Foundation under Grant No. 9909875. Any opinions, findings, and conclusions or recommendations expressed in this material are those of the authors and do not necessarily reflect the views of the funders.

\bibliography{mybibfile}

\noindent\parbox{11.3cm}{\parpic{\includegraphics[width=30mm,scale=0.1]{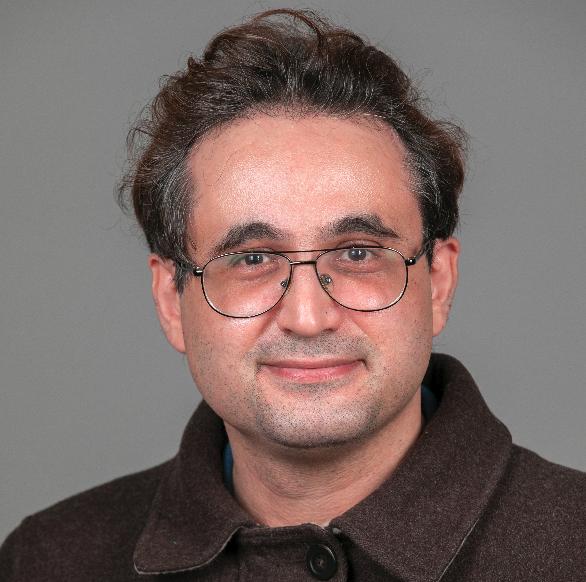}}{\small\quad {\bf Matthew {Korban}} received his BSc and MSc degree in Electrical Engineering in 2013 from the University of Guilan, where he worked on sign language recognition in video. He received his PhD in Computer Engineering from Louisiana State University. He is currently a Postdoc Research Associate at the University of Virginia, working with Prof. Scott T. Acton. His research interest includes Human Action Recognition, Early Action Recognition, Motion Synthesis, and Human Geometric Modeling in Virtual Reality environments.}\\[1mm]}

\noindent\parbox{11.3cm}{\parpic{\includegraphics[width=30mm,scale=0.1]{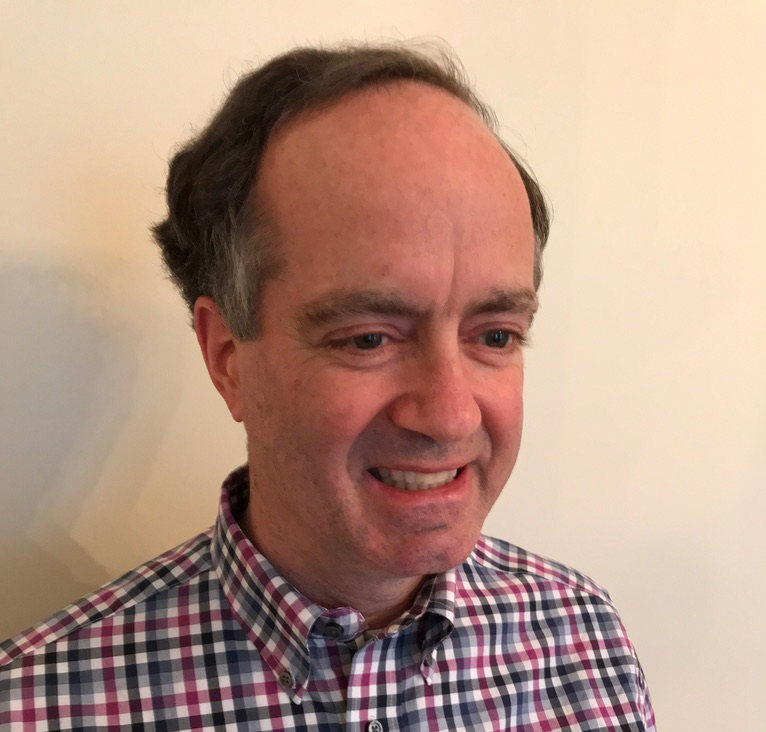}}{\small\quad {\bf Peter {Youngs}} is a professor in the Department of Curriculum, Instruction and Special Education at the University of Virginia. He studies how neural networks can be used to automatically classify instructional activities in videos of elementary mathematics and reading instruction. He currently serves as co-editor of American Educational Research Journal. }\\[1mm]}

\noindent\parbox{11.3cm}{\parpic{\includegraphics[width=30mm,scale=0.1]{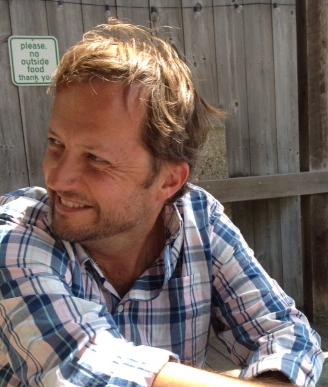}}{\small\quad {\bf Scott T. {Acton}} received his M.S. and Ph.D. degrees at the University of Texas at Austin. He received his B.S. degree at Virginia Tech. Professor Acton is a Fellow of the IEEE ``for contributions to biomedical image analysis.”
Currently, Acton is a program director in the Computer and Information Science and Engineering at the U.S. National Science Foundation. He is also professor of Electrical and Computer Engineering and of Biomedical Engineering at the University of Virginia. Professor Acton’s laboratory at UVA is called VIVA - Virginia Image and Video Analysis. They specialize in bioimage analysis problems.  Professor Acton has over 300 publications in the image analysis area including the books \textit{Biomedical Image Analysis: Tracking} and \textit{Biomedical Image Analysis: Segmentation}. He was the 2018 Co-Chair of the IEEE International Symposium on Biomedical Imaging. Professor Acton was recently Editor-in-Chief of the IEEE Transactions on Image Processing (2014-2018).}\\[1mm]}

\end{document}